\documentclass[12pt]{article}
\usepackage{graphicx} 
\usepackage[labeled,resetlabels]{multibib}
\usepackage{url}
\usepackage[T1]{fontenc}
\usepackage{titlesec}
\usepackage[hidelinks]{hyperref}

\usepackage{enumitem}
\usepackage{soul}
\usepackage{color, colortbl}
\sethlcolor{yellow}
\titleformat{\section}
{\normalfont\bfseries\MakeUppercase}
{}{0em}{}

\titleformat{\subsection}
{\normalfont\bfseries}
{}{0em}{}

\title{Conversational Health Agents: A Personalized LLM-Powered Agent Framework
}
\author{Mahyar Abbasian, M.Sc.\textsuperscript{1*}, Iman Azimi, Ph.D.\textsuperscript{1},  \\
 Amir M. Rahmani, Ph.D.\textsuperscript{1}, and Ramesh Jain, Ph.D.\textsuperscript{1} ~\\\small\normalfont 
~\\\textsuperscript{1}{University of California, Irvine}
~\\\textsuperscript{*}{Corresponding author, abbasiam@uci.edu}}

\begin{document}
\maketitle

\section*{Abstract}

Conversational Health Agents (CHAs) are interactive systems that provide healthcare services, such as assistance and diagnosis. Current CHAs, especially those utilizing Large Language Models (LLMs), primarily focus on conversation aspects. However, they offer limited agent capabilities, specifically lacking multi-step problem-solving, personalized conversations, and multimodal data analysis.
Our aim is to overcome these limitations.
We propose openCHA, an open-source LLM-powered framework, to empower conversational agents to generate a personalized response for users' healthcare queries. This framework enables developers to integrate external sources including data sources, knowledge bases, and analysis models, into their LLM-based solutions. openCHA includes an orchestrator to plan and execute actions for gathering information from external sources, essential for formulating responses to user inquiries. It facilitates knowledge acquisition, problem-solving capabilities, multilingual and multimodal conversations, and fosters interaction with various AI platforms.
We illustrate the framework's proficiency in handling complex healthcare tasks via two demonstrations and four use cases. Moreover, we release openCHA as open source available to the community via GitHub\footnote{\url{https://github.com/Institute4FutureHealth/CHA}}\textsuperscript{,}\footnote{\url{https://opencha.com}}.

\section{Introduction}
Artificial intelligence (AI), particularly large language model (LLM)-based conversational systems, has attracted immense global attention in recent years. These systems have revolutionized the field by enabling unprecedented access to and interaction with vast amounts of textual information. LLMs can aggregate and process comprehensive or focused segments of textual knowledge existing online, delivering contextually relevant, goal-oriented, and interactive access to this knowledge for anyone who needs it. The advent of LLMs has transformed early, simple conversation systems like Alexa and Siri, demonstrating significant effectiveness across diverse domains \cite{chatgptwebsite, han2023medalpaca, tu2023towards}. Conversational systems can now engage in open-ended conversations and provide relevant, contextual information in a more natural and engaging way.

While the field of AI has long explored intelligent agents, their focus has primarily been on analyzing the environment and making decisions based on gathered information. Early AI research often concentrated on physical world problems, fueled by advancements in computer vision, audio processing, and other areas of multimodal perceptual understanding. However, in dynamic environments like health management, where personalized and constantly evolving human health states are crucial, intelligent agents need to accurately capture these states through various means, including conversational interactions and access to personal user data. This information needs to be collected and analyzed, leveraging the vast knowledge gathered through research and practitioners' experience.

Conversational Health Agents (CHAs) hold significant potential to address the challenges of dynamic health management environments. Thanks to the emergence of LLMs, CHAs can now understand user interactions through multimodal conversations, encompassing text, speech, and potentially other modalities. By analyzing these interactions, CHAs need to identify the necessary data, information, computational processes, and knowledge sources required to comprehend the user's evolving health state. This information is then translated into actionable insights that effectively guide healthcare management. In essence, CHAs should combine the power of LLM-based conversions with agents' capabilities, leveraging external data and information sources to navigate the complexities of personalized health environments and provide customized support for users:
\begin{itemize}
    \item \textit{Conversation} is the fundamental mode of human interaction. Throughout the ages, conversations have consistently served as the primary source of knowledge and the catalyst for societal actions. Recently, numerous studies have substantiated the efficacy, usability, and overall satisfaction associated with the conversational aspect of CHAs \cite{milne2020effectiveness}. In healthcare settings, \textit{empathy} \cite{moudatsou2020role, abbasian2023foundation} and \textit{companionship} \cite{moreira2023multichannel, a16z} necessitate personalized conversations.
    \item \textit{Agents} should be furnished with conversational tools, interfaces, computational capabilities, and access to external resources to enhance the quality of healthcare delivery \cite{denecke2023investigating}. Given the intricacies of the healthcare domain \cite{plsek2001challenge, sturmberg2009complexity}, agents should understand users' queries, decompose them into the components of knowledge possession, perform health data access and analysis, and apply reasoning to respond effectively to specific situations \cite{sturmberg2009complexity}. Their adaptability is essential, aligning their evolution with advancements in healthcare technology and literature \cite{plsek2001challenge}. Therefore, agents empower the abilities for \textit{personalization}, \textit{multimodality}, and continuous \textit{up-to-dateness} \cite{abbasian2023foundation}.
\end{itemize}

\begin{figure}[!t]
\centering
\includegraphics[width=\textwidth, trim={2.5cm 6.5cm 18cm 5cm},clip]{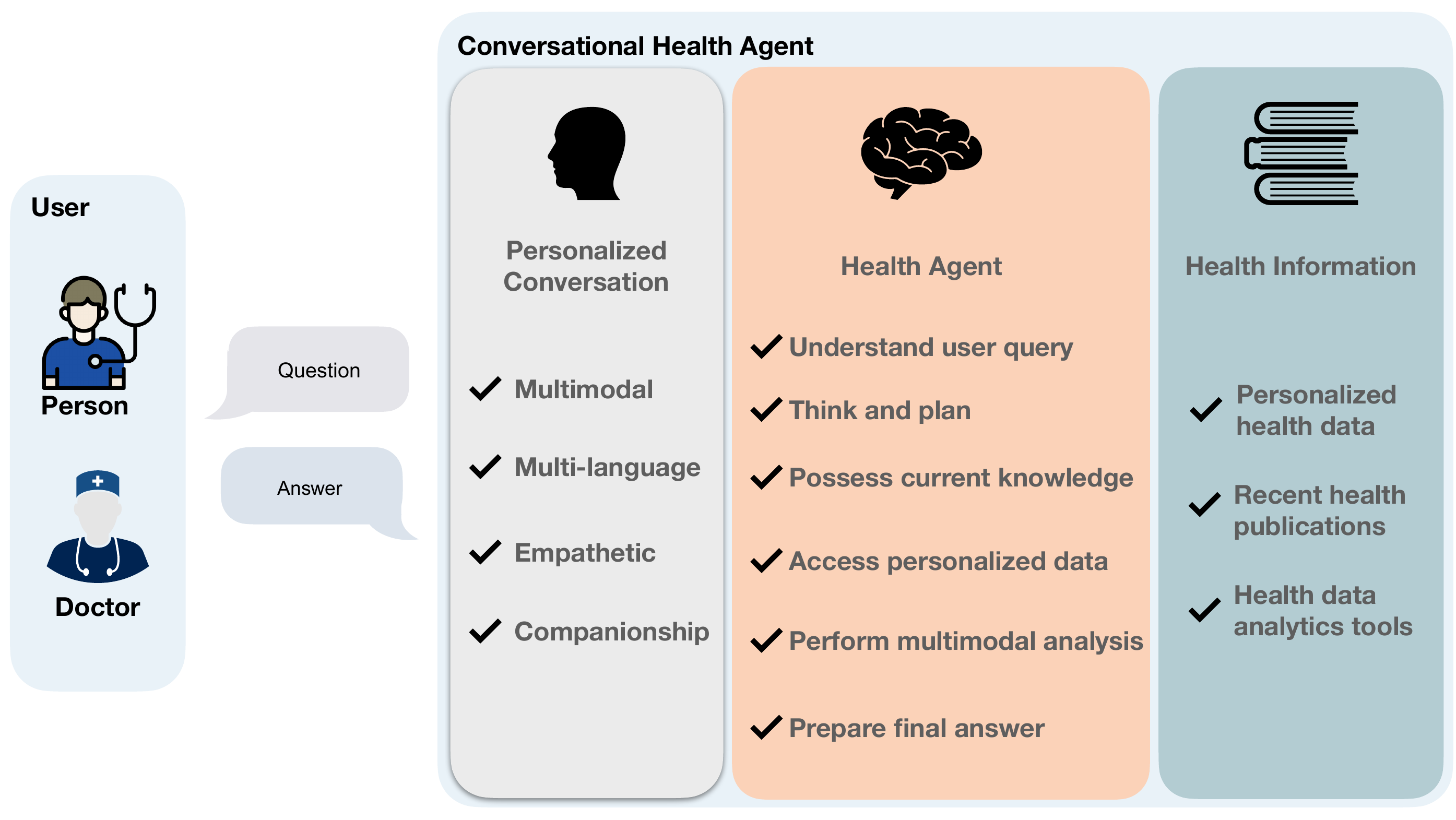}
\caption {A conversational health agent including 1) a conversation component to enable user interaction and 2) a health agent for problem-solving and determining the optimal sequences of actions, leveraging health information.
\label{fig:cha}}
\end{figure}

Our current exploration centers on the development of CHAs using the latest technological developments in AI, LLMs, and mHealth, where it has shown efficacy in the continuous collection of lifestyle and physiological data from users. Figure~\ref{fig:cha} shows an overview of the CHA main components. These indispensable components stand poised to facilitate the creation of exceptionally efficient CHAs.

Existing LLMs, such as ChatGPT \cite{chatgptwebsite}, BioGPT \cite{luo2022biogpt}, ChatDoctor \cite{li2023chatdoctor}, and Med-PaLM \cite{tu2023towards}, are currently active in the medical knowledge domain. These LLMs can be served as CHAs \cite{meier2019feelfit, montenegro2019survey, denecke2023investigating, tudor2020conversational, laranjo2018conversational}. However, they merely focus on the conversational aspects, offering limited agent capabilities such as basic text-based chat interfaces and lacking multi-step problem-solving capabilities. They lack access to users' personally collected longitudinal data and electronic health records (EHRs), which include crucial information like vital signs, biosignals (e.g., electrocardiogram), medical images, and demographic data. Consequently, their responses tend to be generic and may not address individual health circumstances adequately. Moreover, they struggle to incorporate the latest health insights, leading to potentially outdated responses  \cite{zhao2023survey}. Furthermore, these chatbots do not seamlessly integrate with established and existing AI models and tools \cite{han2023randomized} for multimodal predictive modeling, rendering previous healthcare efforts obsolete.

In light of the significant advancements in technology and its paramount importance for both humanity and the environment, it becomes imperative that we synergize all available tools and harness knowledge from diverse sources to craft CHAs that offer a trustworthy, understandable, and actionable environment for a global audience. Presently, we stand on the threshold of crafting frameworks capable of delivering information in the most user-friendly and culturally attuned manner possible. This paper aims to introduce an initial iteration of such agents and lay the foundation for developing more sophisticated tools as our journey unfolds.

\section{Background and Significance}
\subsection{Related Work}
Efforts in developing LLM-based CHAs can be categorized into three main groups: LLM Chatbots, Specialized Health LLMs, and Multimodal Health LLMs. LLM Chatbots employ and evaluate current chatbots (e.g., ChatGPT) in executing distinct healthcare functions \cite{sallam2023chatgpt, biswas2023chatgpt, aydin2022openai, chen2023utility, liu2023biosignal}. For instance, Chen et al. \cite{chen2023utility} examined ChatGPT's efficacy in furnishing dependable insights on cancer treatment-related inquiries. 

Specialized Health LLMs delved deeper into the fundamental aspects of LLMs, aiming to enhance conversational models' performance by creating entirely new LLMs pretrained specifically for healthcare or fine-tuning existing models. Notable examples include initiatives such as ChatDoctor \cite{li2023chatdoctor}, MedAlpaca \cite{han2023medalpaca}, and BioGPT \cite{luo2022biogpt}. This category emerged in response to research indicating that general-domain LLMs often struggle with healthcare-specific tasks due to domain shift \cite{lee2020biobert, gu2021domain}, and relying solely on prompt engineering may not significantly improve their healthcare-specific performance \cite{gutierrez2022thinking, moradi2021gpt}.

Multimodal Health LLMs involve a novel trajectory by integrating multimodality into LLMs for diagnostic functions. For instance, Tu et al. \cite{tu2023towards} investigated the potential of foundational transformer concepts in LLMs to amalgamate diverse modalities—videos, images, signals, and text—culminating in a multimodal generative model. Xu et al. \cite{xu2023elixr} introduced an LLM-aligned multimodal model, coupling chest X-ray images with radiology reports for X-ray-related tasks. Similarly, Belyaeva et al. \cite{belyaeva2023multimodal} incorporated tabular health data into LLMs, yielding multimodal healthcare capabilities.

\subsection{Existing Research Gaps and Challenges}
\textit{Knowledge-groundedness} and \textit{personalization} in CHAs require tailored interactions that transcend basic dialogues, ensuring inclusivity through versatile, multimodal multilingual interfaces. The goal is to create CHAs that not only excel in conversational skills but also exhibit agent capabilities, enabling them to engage in critical thinking and strategic planning as proficient problem solvers. 
Despite the great efforts in developing CHAs, the existing services and models suffer from the following limitations:
\begin{enumerate}[label=\roman*)]
    \item \ul{Insufficient support for comprehensive personalization, particularly in cases necessitating real-time access to individualized data.} A substantial portion of users' healthcare data, primarily images, time-series, tabular data, and all other users' measured personal data streams is housed within healthcare platforms. Currently, CHAs have limited access to this data, primarily during the training and fine-tuning phases of LLM development, or they are completely severed from user data thereafter. The absence of accurate user healthcare information -- including continuous data from wearable devices, mHealth applications, and similar sources -- hampers the performance of these agents, confining their capabilities to furnish generic responses, offer general guidelines, or potentially provide inaccurate answers.
    \item \ul{Limited capacity to access up-to-date knowledge and retrieve the most recent healthcare knowledge base.} Conventional LLMs depend on limited data and Internet-derived knowledge during their training phase, leading to three primary challenges. They tend to exhibit biases favoring populations with the most abundant online content, underscoring the importance of accessing the latest, relevant data. Recently introduced LLM-based services (e.g., ChatGPT4 \cite{chatgptwebsite}) offer Internet search, but this is still insufficient for healthcare applications due to the large number of websites propagating false information. They lack updates on newly reliable Internet resources, a critical shortcoming in healthcare where novel, reliable, and evaluated treatments and modifications to previous recommendations are frequently ignored. Lastly, their reliance on outdated or less pertinent data makes identifying instances of hallucination \cite{guerreiro2023hallucinations} problematic. Lack of up-to-date information reduces the trustworthiness and credibility of generated responses \cite{zhao2023survey}. 
    \item \ul{Lack of seamless integration with established, multimodal data analysis tools and predictive models that require external execution.} Current agents often overestimate the computational capabilities of generative AI, leading to an under-utilization of well-established healthcare analysis tools, despite their proficiency in managing diverse data types \cite{han2023randomized}. 
    \item \ul{Lack of multi-step problem-solving capabilities.} Existing LLM-based CHAs are typically specialized for specific tasks or deficient in robust data analysis capabilities. For example, Xu et al. \cite{xu2023elixr} model performs X-ray image reporting relying solely on X-ray images, ignoring other modalities, such as vital signs recorded in time-series format. Additionally, the existing CHAs cannot address intricate sequential tasks (i.e., act as problem solvers). Incorporating LLMs into CHAs requires integrating sequential reasoning, personalized health history analysis, and data fusion.
\end{enumerate}

LLMs solely are insufficient to tackle the previously mentioned challenges. To make them practical for real-world applications, we need a comprehensive framework that harnesses LLMs while integrating various auxiliary components and external resources. 

\subsection{Key Contributions}
In this article, we present a holistic LLM-powered framework for the development of CHA, aiming to rectify the limitations mentioned above. We delineate the components of our framework and proceed with a case study demonstrating our agent's capacity. The framework is a problem solver that provides personalized responses by utilizing contemporary Internet resources and advanced multimodal healthcare analysis tools, including machine learning methods. Significant contributions are as follows.

\begin{enumerate}

\item We propose an LLM-powered Orchestrator, acting as a problem solver, to address healthcare-related queries by analyzing input queries, gathering the required information, performing actions, and offering personalized responses.

\item We introduce external healthcare data sources, knowledge bases, and AI and Analysis models critical for enabling CHAs to offer reliable, trustworthy, and up-to-date responses.

\item We incorporate multimodal and multilingual capabilities into the framework, increasing usability.

\item We show the framework's effectiveness using two demos and four use cases.

\item We release the CHA framework as open source on GitHub\footnote{~\url{https://github.com/Institute4FutureHealth/CHA}} along with detailed documentation\footnote{~\url{https://docs.opencha.com/}}, inviting the community to leverage and integrate it into their solutions.

\end{enumerate}

\section{Material and Methods}
\begin{figure*}[t]
\includegraphics[width=\textwidth, trim={2.3cm 0.8cm 0.2cm 0cm},clip]{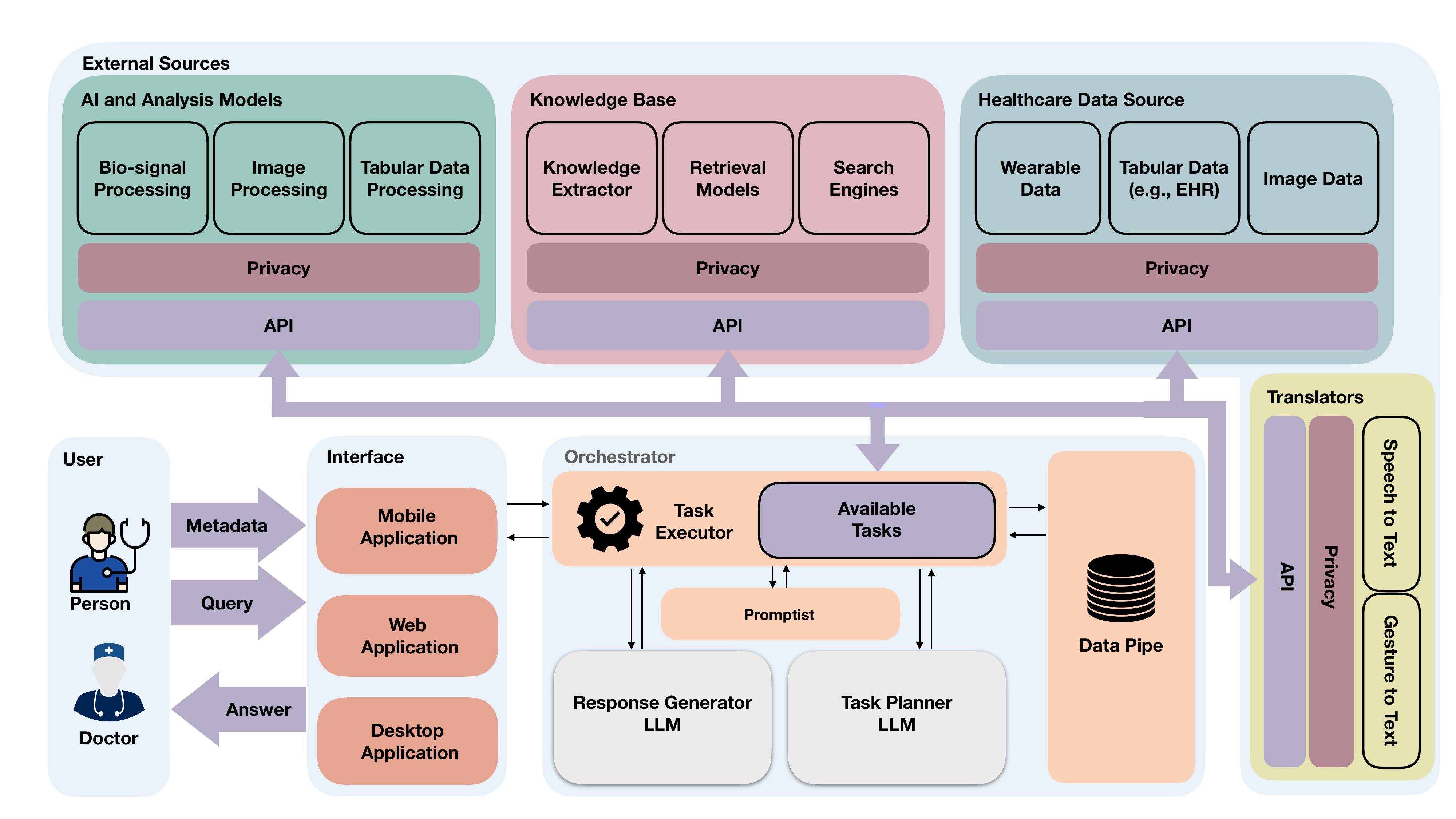}
\centering
\caption {An overview of the proposed LLM-powered framework leveraging a service-based architecture} \label{fig:architecture}
\end{figure*}

We design an LLM-powered framework with a central agent that perceives and analyzes user queries, provides appropriate responses, and manages access to external resources through Application Program Interfaces (APIs) or function calls. The user-framework interaction is bidirectional, ensuring a conversational tone for ongoing and follow-up conversations. Figure \ref{fig:architecture} shows an overview of the framework, including three major components:  \textit{Interface}, \textit{Orchestrator}, and \textit{External Sources}.

\subsection{Interface}
\textit{Interface} acts as a bridge between the users and agents, including interactive tools accessible through mobile, desktop, or web applications. It integrates multimodal communication channels, such as text and audio. The Interface receives users' queries and subsequently transmits them to the \textit{Orchestrator} (see Figure \ref{fig:architecture}). 

Within this framework, users can provide metadata (alongside their queries), including images, audio, gestures, and more. For instance, a user could capture an image of their meal and inquire about its nutritional values or calorie content, with the image serving as metadata.

\subsection{Orchestrator}

The \textit{Orchestrator} is the openCHA agent core, which is responsible for problem-solving, planning, executing actions, and providing an appropriate response based on the user query. It incorporates the concept of the Perceptual Cycle Model \cite{neisser1978perceiving} in openCHA, allowing it to perceive, transform, and analyze the world (i.e., input query and metadata) to generate appropriate responses. To this end, the input data are aggregated, transformed into structured data, and then analyzed to plan and execute actions. Through this process, the Orchestrator interacts with external sources to acquire the required information, perform data integration and analysis, 
 and extract insights, among other functions. In the following, we outline five major components of the Orchestrator.

The \textit{Task Planner} is the LLM-enabled decision-making, planning, and reasoning core of the Orchestrator. Its primary responsibility is gathering all necessary information to answer users' queries. To achieve this, it interprets the user's query and metadata, identifying the necessary steps for task execution. 

To transform a user query into a sequence of tasks, we incorporate the Tree of Thought \cite{yao2023tree} prompting methods into the Task Planner. Using this prompting method, the LLM is asked to 1) generate three unique strategies (i.e., sequences of tasks to be called with their inputs), 2) describe the pros and cons of each strategy, and 3) select one as the best strategy. An alternative prompting technique incorporated in openCHA is ReAct \cite{yao2022react}, which employs reasoning and action techniques to ascertain the essential tasks to be executed. openCHA offers users the flexibility to choose the prompting method that best meets the needs of their application. Other prompting techniques, such as Plan-and-Solve Prompting \cite{wang2023plan}, could also be implemented and integrated as a \textit{Task Planner}. 

We outline the process of creating and integrating a task into openCHA in Appendix 1. We also indicate how the task is converted into an appropriate prompt, enabling the \textit{Task Planner} to recognize the available tasks and how to invoke them. Appendix 2 provides a detailed examination of the \textit{Task Planner's} implementation, utilizing the Tree of Thought prompting method.

In the proposed Orchestrator, the planning part is performed in English, leveraging the superior capabilities of LLMs in this language. The framework can employ one of two distinct approaches if the query is in a language other than English. The first approach retrains the source language and utilizes the language model capabilities in that language to generate responses. The second approach involves translating the query into English (e.g., using Google Translate), planning and executing the process in English, and translating the final answer back into the source language. 

The \textit{Task Executor} carries out actuation within the Orchestrator by following the planning and task execution steps determined by the \textit{Task Planner}. The Task Executor has two primary responsibilities. First, it acts as a data converter, converting the input query and metadata and preparing it to be used by the \textit{Task Planner}.
For instance, if the question is in a language other than English, it will be translated into English using the Google Translate service \cite{translator}. Furthermore, if the metadata contains files or images, \textit{Task Executor} sends the metadata details to \textit{Task Planner} for planning.
Second, the \textit{Task Executor} executes tasks generated by the \textit{Task Planner} through interactions with external sources. The results are then relayed to the \textit{Task Planner} to continue planning if needed. In the end, the \textit{Task Planner} signals the end of the planning. In Appendix 2, we detail how the \textit{Task Planner} translates planned tasks into execution instructions, enabling the \textit{Task Executor} to properly carry out the tasks.

It is crucial to emphasize that communication between the task planner and task executor is bidirectional. An iterative process continues between the Task Executor and Task Planner until the Task Planner accumulates sufficient information to respond appropriately to the user's inquiry. This two-way exchange proves indispensable because, in specific scenarios, the Task Planner may necessitate intermediate information to determine subsequent actions.

The \textit{Data Pipe} is a repository of metadata and data acquired from \textit{External Sources} through the execution of conversational sessions. This component is essential because numerous multimodal analyses involve intermediate stages, and their associated data must be retained for future retrieval. The intermediate data might be large, surpassing token limits, or challenging to comprehend and utilize by the \textit{Task Planner's} or \textit{Response Generator's} LLM. The \textit{Data Pipe} is automatically managed by the \textit{Task Executor}. 
It monitors the stored metadata and intermediate data. 

The \textit{Data Pipe} in openCHA can range from a simple in-memory key/value storage for intermediate data to a more complex database system. The proposed framework allows developers to determine whether their tasks' results are intermediate or should be directly returned to the LLM. Appendix 1 details how developers can configure this setting.

Additionally, we have implemented a mechanism whereby an intermediate result stored in the \textit{Data Pipe} generates a unique key as the task's outcome. This key is then provided in the \textit{Task Planner} prompt, aiding the \textit{Task Planner} in recognizing and utilizing this data as necessary. Appendix 4 illustrates sample prompts generated for tasks and demonstrates how the \textit{Task Planner} employs the \textit{Data Pipe} key.

The \textit{Promptist} is responsible for transforming query text or outcomes from External Sources into suitable prompts that can be supplied to either the \textit{Task Planner} or the \textit{Response Generator}. The \textit{Promptist} provides the flexibility to modify and adapt each technique, allowing for seamless integration and customization. It can be implemented using existing prompting techniques, some of which are listed as follows. 

LLM-REC, proposed by Lyu et al. \cite{lyu2023llm}, employs four unique prompting strategies to enrich text descriptions, enhancing personalized text-based recommendations. The approach leverages the LLM to understand item characteristics, significantly improving recommendation quality. Additionally, the Hao et al. \cite{hao2022optimizing} method can be leveraged, which optimizes text-to-image prompt generation through a framework called prompt adaptation. It automatically refines user inputs into model-preferred prompts. This process starts with supervised fine-tuning of a pretrained language model using a curated set of prompts. It then employs reinforcement learning, guided by a reward function, to identify more effective prompts that produce aesthetically pleasing images aligned with user intentions. Furthermore, the instructions provided by OpenAI on creating more effective prompts can be used \cite{chatgptpromptwebsite}.


The \textit{Response Generator} is an LLM-based module responsible for preparing the response. It refines the gathered information by the \textit{Task Planner}, converting it into an understandable format and inferring the appropriate response. We separate the \textit{Response Generator} and \textit{Task Planner} to allow flexibility in choosing diverse LLM models and prompting techniques for these components. This division ensures that the \textit{Task Planner} focuses solely on planning without responding to users, while the \textit{Response Generator} utilizes gathered information to deliver conclusive responses. This segregation facilitates the \textit{Response Generator} in addressing aspects of \textit{empathy} and \textit{companionship} in conversations. In contrast, the \textit{Task Planner} primarily handles \textit{personalization} and the \textit{up-to-dateness} of conversations. Appendix 3 outlines the implementation of the \textit{Response Generator} and how it utilizes results collected by the \textit{Task Planner} to respond to the user effectively.

\subsection{External Sources}
\textit{External Sources} play a pivotal role in obtaining essential information from the broader world. Typically, these External Sources furnish application program interfaces (APIs) that the Orchestrator can use to retrieve required data, process them using AI or analysis tools, and extract meaningful health information. 
In openCHA, we integrate with four primary external sources, which we found critical for CHAs (see Figure \ref{fig:architecture}).

\textit{Healthcare Data Source} enables the collection, ingestion, and integration of data captured from a variety of sources, such as Electronic Health Record (EHR), smartphones, and smartwatches, for healthcare purposes \cite{labbaf2023zotcare}. Examples of data sources are mHealth platforms and healthcare databases. mHealth platforms have garnered significant attention in the recent wave of healthcare digitalization, enabling ubiquitous health monitoring \cite{jusoh2017survey,chiarini2013mhealth}. The data encompass various modalities, including biosignals (e.g., PPG collected via a smartwatch), images (e.g., captured via user's smartphone), videos, tabular data (e.g., demographic data gathered from EHR), and more.
Notable examples of such healthcare platforms include ZotCare \cite{labbaf5zotcare} and ilumivu \cite{ilumivu}, offering APIs for third-party integration. In our context, the Orchestrator functions as a third party, accessing user data with their consent.

\textit{Knowledge Base} fetches the most current and pertinent data from healthcare sources, such as healthcare literature, reputable websites, or knowledge graphs using search engines or retrieval models \cite{nori2023can, hiemstra2009information, turtle1992comparison, norican}. Accessing this retrieved information equips CHAs with up-to-date, personalized knowledge, enhancing its trustworthiness while reducing hallucination and bias. openCHA allows the integration of various knowledge bases to be defined and configured as tasks.

\textit{AI and Analysis Models} provide data analytics tools to extract information, associations, and insights from data \cite{topol2023artificial, han2023randomized}, playing a crucial role in the evolving landscape of LLM-healthcare integration, enhancing trustworthiness and personalization. They can perform various tasks, including data denoising, abstraction, classification, and event detection, to mention a few \cite{topol2023artificial, han2023randomized}. As generative models, LLMs cannot effectively perform extensive computations or act as machine learning inferences on data. The AI platforms empower our framework to leverage existing health data analytic approaches.

\textit{Translators} effectively convert various languages into widely spoken languages, such as English, thereby enhancing the accessibility and inclusivity of CHAs. Existing agents face limitations that hinder their usability for large communities globally. Universal text literacy for CHAs often narrows their reach and positions them as a privilege \cite{clarke2016health, jain2003folk, borovoy2002folk}. Many underserved communities face obstacles while using CHAs due to their educational disparities, financial constraints, and biases that favor developed nations within existing technological paradigms. Our framework integrates with Translator platforms and is designed to accommodate and support communication with diverse communities. This integration enhances the overall usability of CHAs. 

The selection of the external sources is based on the information and knowledge they provide and their interaction with the Orchestrator. The Orchestrator handles various data types in the proposed framework, including text, JSON formatted data, or unstructured data such as images and audio. This design ensures that any external source capable of returning results in these formats is supported.

\section{Demonstration}
\begin{figure*}[!t]
\vspace*{-1.7cm}
\includegraphics[width=\textwidth]{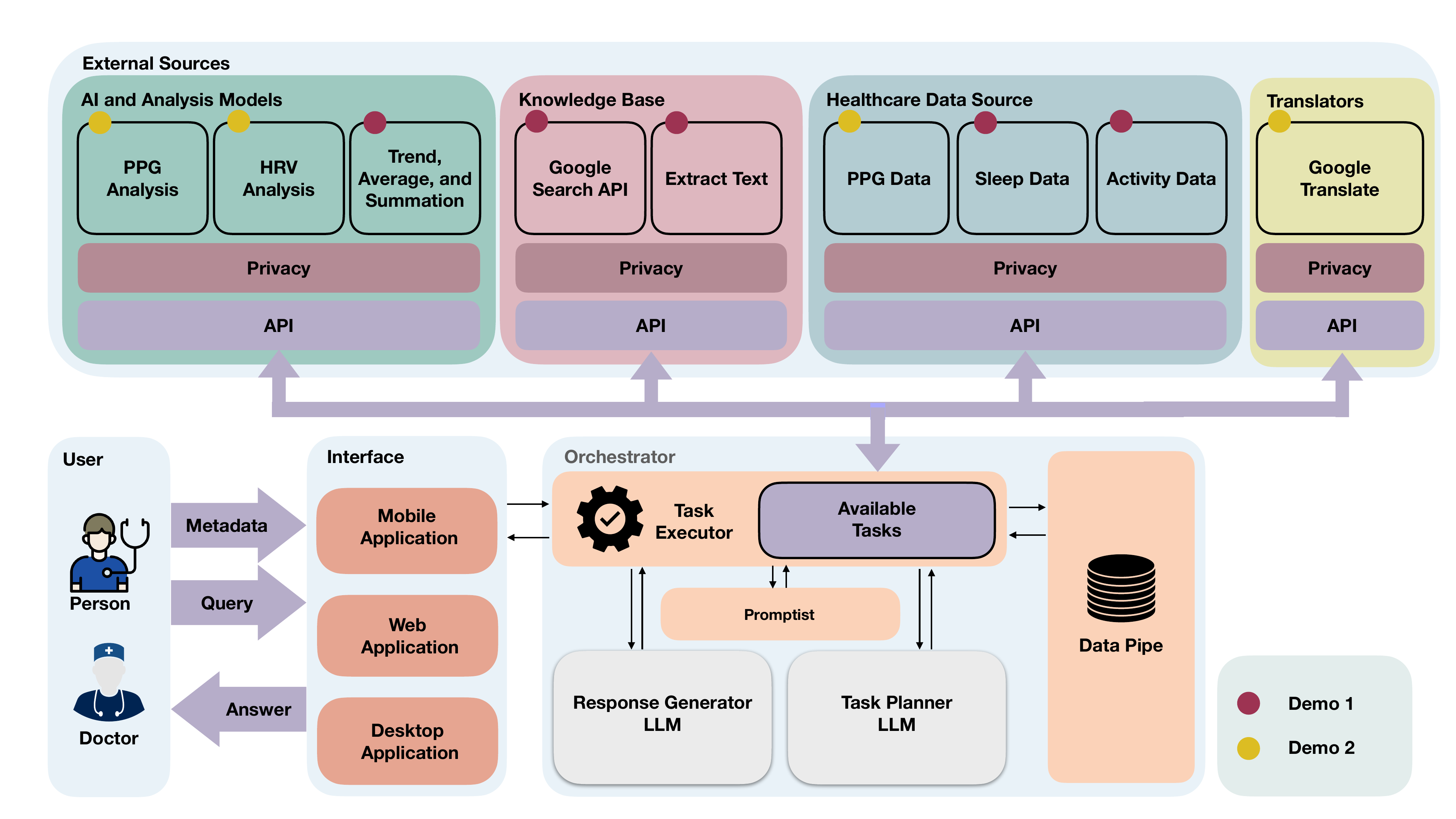}\hfill
\vspace*{-0.06cm}
\centering
\caption {Overview of the implemented tasks and components and how they are used in the two demos.} \label{fig:implementation}
\end{figure*}

We demonstrate the capabilities of openCHA through three distinct demos. These demonstrations highlight how LLMs' planning and reasoning abilities can effectively comprehend user queries and translate them into appropriate task executions. Each demonstration involves linking a set of implemented tasks to openCHA. Subsequently, we highlight openCHA's planning proficiency in effectively sequencing tasks in the correct order with appropriate inputs and executing the tasks. After executing all necessary tasks, the results are conveyed to the Response Generator to provide the final response. The overview of the implemented tasks and their utilization in each demo is depicted in Figure \ref{fig:implementation}.

\subsection{Demo 1: Patient health record reporting}

The first demo indicates how openCHA interacts with patient data stored in a database, conducts data analysis, generates health reports upon request, and asks follow-up questions. Examples of the user's questions could be: "Provide a sleep summary of Patient 5 during August 2020", "Is Patient 5 REM sleep enough during August 2020?", "How much is the total step count of Patient 5 during August 2020?" or "Provide an activity summary of Patient 5 during 2020". For this demo, we implement two tasks for retrieving sleep and physical activity from a health monitoring dataset \cite{labbaf2023physiological}. The data utilized in this demo is a part of an extensive longitudinal study focusing on the mental health of college students, as documented in \cite{labbaf2023physiological}. Moreover, we develop analytical tasks capable of executing basic statistical analysis (e.g., computing trends and averaging). Finally, we also add Google Search and Extract Text tasks so that the current setup of openCHA can access internet information. The involved tasks in this demo are marked by red dots in Figure \ref{fig:implementation}.

\begin{figure*}[!t]
\includegraphics[width=\textwidth, trim={0cm 15.5cm 0cm 0.9cm},clip]{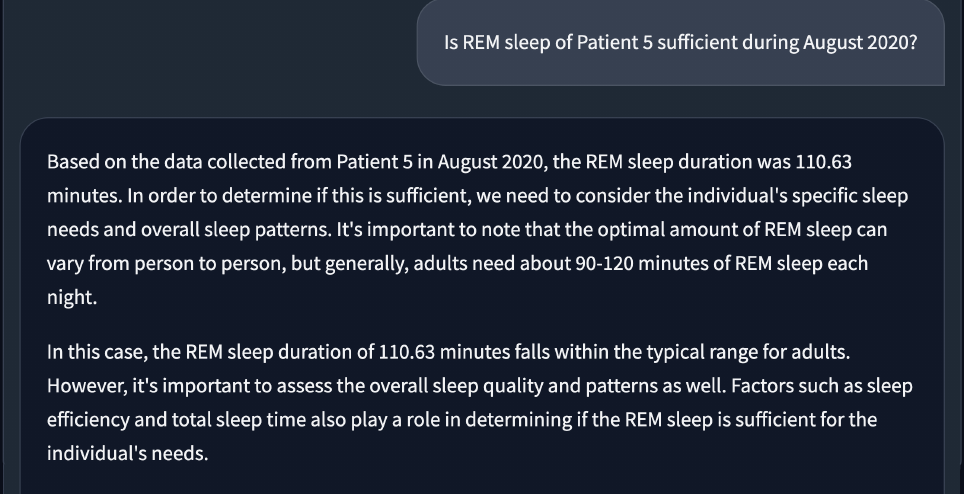}\hfill
\vspace*{-0.06cm}
\includegraphics[width=\textwidth, trim={0cm 1cm 0cm 14cm},clip]{figs/scenario2_1.pdf}\hfill
\vspace*{-0.09cm}
\includegraphics[width=\textwidth, trim={0cm 0.5cm 0cm 1cm},clip]{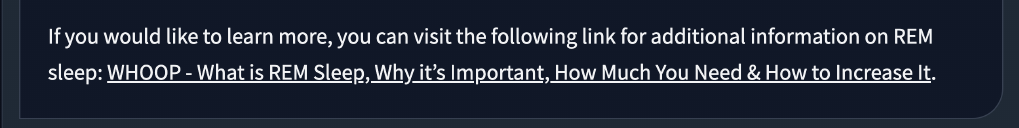}\hfill
\centering
\caption {Demo 1. Patient health record reporting and follow up.} \label{fig:scenario2}
\end{figure*}

Figure \ref{fig:scenario2} illustrates sample queries and the corresponding responses from openCHA. The Task Planner identifies the correct tasks, their execution order, and the appropriate inputs to be used. Table \ref{tab:scenario2} showcases the strategy chosen by the Task Planner using the Tree of Thought prompting technique.

\begin{table*}[t]
    \centering
    \caption{The strategies suggested by Tree of Thought prompting technique for the first question.}
    \resizebox{\textwidth}{!}{
    \begin{tabular}{|c|l|} \hline 
         & I will proceed to directly analyze the REM sleep data of\\
         & Patient 5 for August 2020, providing a precise and specific\\ 
         &  conclusion. \\
         & \\
        Decision & Execution: \\
        & 1. Use the sleep\_get tool to obtain the REM sleep data for \\
        & Patient 5 in August 2020. \\
        & 2. Utilize the sleep\_analysis tool to analyze the REM sleep  \\
        & duration and efficiency for August 2020.\\ \hline
    \end{tabular}
    }
    
    \label{tab:scenario2}
\end{table*}

\subsection{Demo 2: Objective stress level estimation with multilanguage interaction}

Demo 2 indicates the openCHA's capability to conduct signal processing and objective stress level estimation. We also show that openCHA can answer the query in multiple languages. This is achieved by interacting with a translator, health data sources, and AI models. Examples of interactions include inquiries such as "Retrieve the stress level of Patient 5 on August 29th, 2020" and "What is the average heart rate of Patient 5 during August 2020?" To fulfill our objective, we implemented three distinct tasks (yellow dots in Figure \ref{fig:implementation}). The first task involved acquiring Photoplethysmogram (PPG) data from the patients. PPG data were gathered using Samsung Gear Sport smartwatches \cite{samsunggear}, with a sampling frequency of 20 Hz, while participants were in free-living conditions. The data is part of the \cite{labbaf2023physiological}. The second task performs PPG signal processing to extract heart rate variability (HRV) metrics. For this purpose, we utilize the Neurokit \cite{Makowski2021neurokit} Python library. In our case study, we extract a total of 32 HRV parameters, including metrics such as the root mean square of successive differences between normal heartbeats (RMSSD), low-frequency (LF), and high-frequency (HF) values \cite{shaffer2017overview, khatibi2023impact}. The third task estimates stress levels based on HRV using an AI model. Initially, we employed an autoencoder to reduce the dataset's 32 HRV features to 12. Subsequently, a four-layer neural network categorizes the 12 features into five stress levels. The evaluation of the stress estimation model demonstrates an 86\% accuracy rate on a test set.

\begin{figure*}[h]
\includegraphics[width=\textwidth]{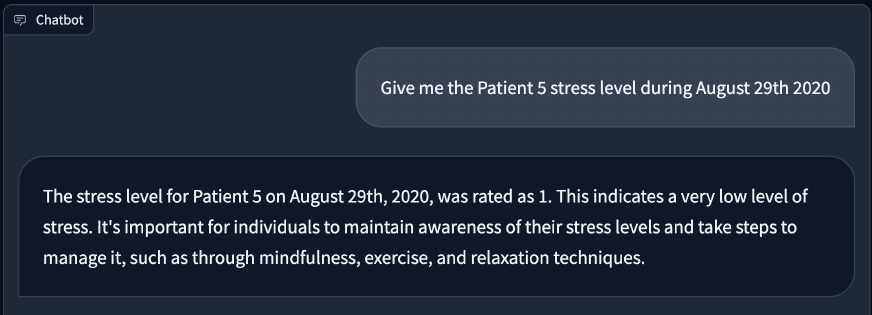}\hfill
\vspace*{-0.09cm}
\includegraphics[width=\textwidth]{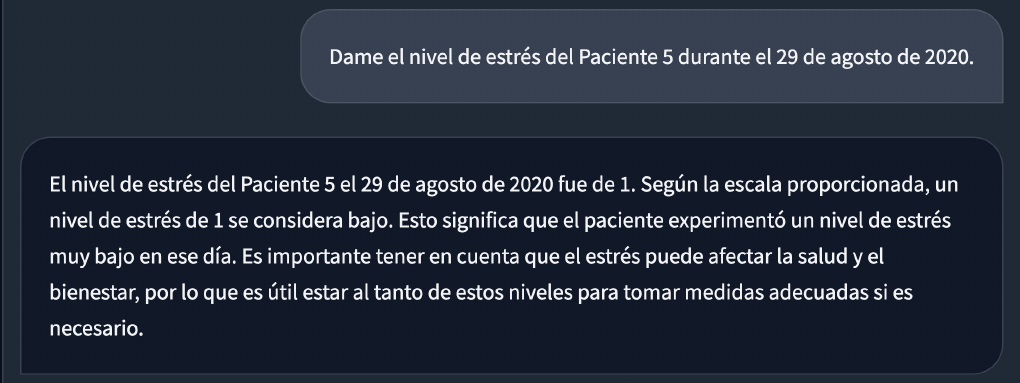}\hfill

\centering
\caption {Demo 2. Objective stress level estimation. The question is asked in English and Spanish.} \label{fig:scenario3}
\end{figure*}

Figure \ref{fig:scenario3} depicts example queries and the corresponding responses from openCHA. The Task Planner's approach involves initially retrieving the PPG data of Patient 5 on August 29th, 2020. Subsequently, the obtained result is forwarded to the PPG analysis task to extract HRV metrics. Lastly, the planner initiates the execution of stress analysis tasks, providing the HRV metrics for this task. Table \ref{tab:scenario3} displays the chosen strategy by the Task Planner utilizing the Tree of Thought prompting technique. The ultimate estimated stress, along with an explanation, is then returned to the user.

\begin{table*}[t]
    \centering
    \caption{The strategies suggested by Tree of Thought prompting technique for the first question.}
    \resizebox{\textwidth}{!}{
    \begin{tabular}{|c|l|} \hline 
        & 
        The best strategy provides both detailed PPG analysis and an \\
        & estimation of the stress level, which offers a comprehensive view \\
        & of the patient's health status. \\
         &
        \\
        & Execution:\\
        Decision &
        1. Use ppg\_get tool to retrieve the PPG data for Patient 5 \\
        &during August 29th, 2020.\\
        & 2. Analyze the PPG data with ppg\_analysis tool to obtain the \\
        & heart rate.\\
        & 3. Use stress\_analysis to estimate the stress level based on the \\
        &obtained PPG data.\\ \hline
    \end{tabular}
    }
    \label{tab:scenario3}
\end{table*}

\subsection{openCHA Use cases}

To indicate the usability of openCHA across various applications, we outline several use cases that have utilized the framework in their research as follows.

\begin{enumerate}

\item ChatDiet \cite{yang2024chatdiet} introduced a personalized, nutrition-oriented food recommendation agent, utilizing openCHA as its core implementation. By integrating personal and population models as external sources, ChatDiet offered tailored food suggestions. It enhanced traditional food recommendation services by delivering dynamic, personalized, and explainable recommendations. In a case study, ChatDiet achieved an effectiveness rate of 92\%, outperforming solutions like ChatGPT.

\item Knowledge-infused LLM-powered CHA for diabetic patients \cite{abbasian2024knowledge} 
 is developed by integrating domain-specific knowledge and analytical tools as external sources using openCHA. This integration included incorporating American Diabetes Association dietary guidelines and deploying analytical tools for nutritional intake calculation, resulting in superior performance compared to GPT4 in managing diabetes through tailored dietary recommendations, as demonstrated by an evaluation of 100 diabetes-related questions.

\item openCHA was employed to develop an agent for evaluating the safety and reliability of mental health chatbots \cite{park2024building}. This agent's evaluation capabilities were compared with expert assessments and several existing LLMs, including GPT4, Claude, Gemini, and Mistral. Guidelines and benchmarks introduced by experts and Internet search served as external sources linked to openCHA. The agent demonstrated superior accuracy, achieving the lowest mean absolute error (MAE) against experts' scores — a reduction by a factor of 1 compared to LLMs' scores, with the maximum MAE being 10 — and provided unbiased evaluation scores.

\item The Empathy-enhanced CHA \cite{abbasian2024empathy} was developed to interpret and respond to users' emotional states through multimodal dialogue, representing a significant step forward in providing contextually aware and empathetically resonant support in the mental health field. This paper utilized speech-to-text, text-to-speech, and speech emotion detection models as external sources connected to openCHA.
\end{enumerate}

To see more demonstrations on how the openCHA works in real setup, we have uploaded multiple YouTube videos\footnote{\url{https://www.youtube.com/watch?v=w48sPlF5zhs}}\textsuperscript{,} \footnote{\url{https://www.youtube.com/watch?v=PWxL_OgWGfE&t=3s}}\textsuperscript{,} \footnote{\url{https://www.youtube.com/watch?v=c-7IEBaRSyQ}}\textsuperscript{,} \footnote{\url{https://www.youtube.com/watch?v=rHXpk_P5n6Y}}

\noindent

\section{Discussion}

\subsection{openCHA Potentials and Limitations}

In this section, we briefly discuss our proposed framework's capabilities, potentials, and limitations.

\textbf{Flexibility:}
openCHA provides a high level of \textit{flexibility} to integrate LLMs with external data sources, knowledge bases, and analytical tools. The proposed components can be developed and replaced according to the requirements of the healthcare application in question. For instance, new external sources can be effortlessly integrated and introduced as new tasks into openCHA. The LLMs employed in openCHA can be readily swapped with fine-tuned or more healthcare-specific LLMs. Similarly, the Planner prompting technique and decision-making processes are modifiable. This flexibility facilitates collaboration among diverse research communities, enabling them to contribute to various aspects of CHAs. Appendix 1 shows how a new task can be defined and introduced into openCHA.

\begin{figure}[t!]
\centering
\includegraphics[width=\textwidth, trim={0cm 0cm 0cm 0cm},clip]{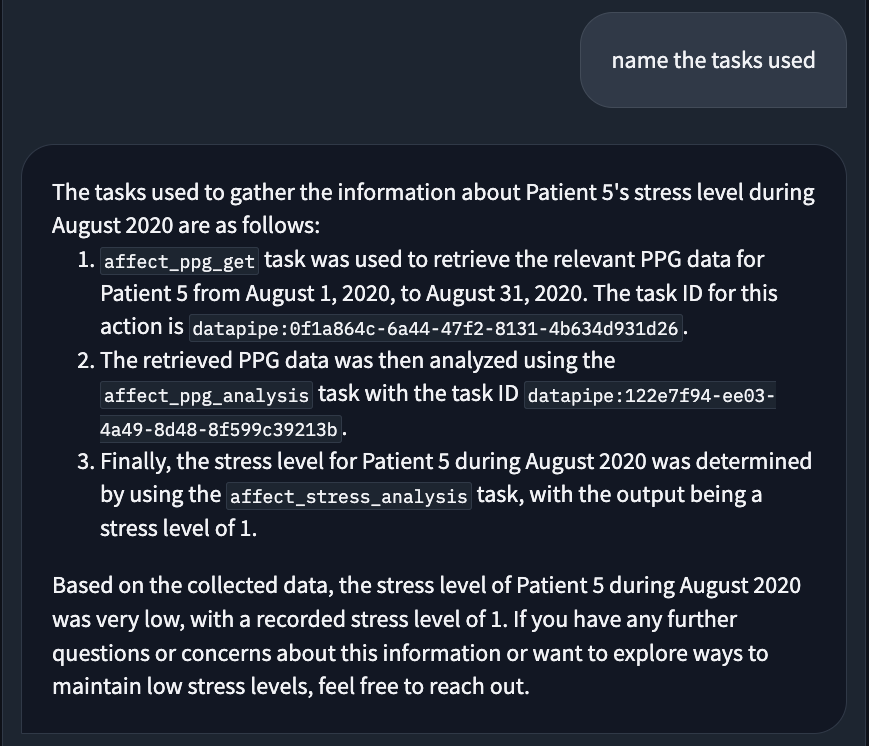}
\caption {A example screenshot indicating how openCHA improves explainability in conversations.
\label{fig:explainability_example}}
\end{figure}

\noindent
\textbf{Explainability:}
openCHA enhances \textit{explainability} for CHAs, allowing users to inquire about the tools and actions used to generate a response. As detailed in Appendices 2 and 3, openCHA maintains a "previous actions" section that records past conversations and tasks. When queried about task usage, it lists the executed tasks and their applications, enhancing transparency and fostering trust between users and CHAs. For instance, in Demo 2, when a user asks, "Name the tasks used," openCHA responds by detailing that PPG and HRV data were utilized to determine stress levels. An example of this interaction is shown in Figure \ref{fig:explainability_example}.

\noindent
\textbf{Personalization:}
The openCHA framework enhances personalization by integrating individual information and analytics tools from healthcare systems or local databases as external sources. The quality of these external sources greatly influences the effectiveness of the personalization. For example, ChatDiet \cite{yang2024chatdiet} utilizes personal dietary preferences and population data, along with an analysis of nutrients' effects on health outcomes like sleep quality, to enhance its food recommendations significantly. This strategy not only heightens the accuracy of the recommendations but also ensures they are precisely tailored to meet individual dietary needs.

\noindent
\textbf{Reliability:} 
openCHA boosts the reliability of answers by leveraging validated information and computations as external sources. Our framework is tailored to effectively utilize existing LLMs for complex healthcare tasks, strategically offloading computational and sensitive information tasks to external sources while reserving LLMs primarily for reasoning and generating responses. For instance, the paper "Knowledge-infused LLM-powered CHA for diabetic patients" \cite{abbasian2024knowledge} demonstrates the benefits of integrating external knowledge to accurately determine nutritional values and align them with established guidelines, highlighting inaccuracies in nutritional estimations when solely relying on GPT4 LLM for data access and calculations.

\noindent
\textbf{Latency:} 
Utilizing multiple external sources offers benefits, but it can also affect the model's response time, potentially leading to increased latency in the CHA. As the number of tasks and steps within the framework expands, there could be a rise in response time, which might diminish usability. Recent research by Singh et al. \cite{singh2024llm} explores new methods for executing tasks in parallel when they are not dependent on each other.

\noindent
\textbf{Toekn Limit:}
Token limits in LLMs present a challenge for accommodating tasks within the Task Planner. However, recent advancements \cite{caludewebsite} indicate progress in extending LLM token limits, which helps mitigate this issue.

\noindent
\textbf{Privacy and Security:}
Privacy and security are crucial in this framework, particularly for healthcare applications dealing with user privacy concerns. Strong privacy measures are essential to prevent unauthorized access, data breaches, and identity theft, with potentially severe consequences \cite{hathaliya2020exhaustive, price2019privacy}. To address privacy and security issues, external sources should provide data confidentiality. For example, this can be enabled by either granting limited access to CHAs as third parties with user permission or employing de-identification and anonymization techniques \cite{xu2014survey}. Another strategy is to prevent LLMs from using users' provided data for training and fine-tuning, ensuring that such data is not stored.


\subsection{Study Limitations and Future Work}

In this section, we outline the study limitations and future research directions.

\noindent
\textbf{Planning Robustness:}
Since we utilize LLMs for planning and response generation, there is still the inherent risk of biases or trustfulness issues. Our framework aims to enhance the robustness of planning by integrating external sources to reduce these problems, though it cannot ensure their complete elimination. To enhance the planning robustness, we will explore using Agentic design patterns like the self-consistency \cite{wang2022self} method or new reasoning techniques.

\noindent
\textbf{Accuracy and Evaluation:}
Accuracy and evaluation in our framework hinge on the configuration choices made by researchers, such as the selected external sources, LLM, and planning technique. Since knowledge, data, and analytics are outsourced to external sources, the quality of these sources plays a crucial role in enhancing accuracy; better external sources increase the likelihood of achieving superior results.

Two distinct assessments are necessary to evaluate such systems. The first evaluates the accuracy of external sources, whether they are AI models or knowledge bases. The second assesses the overall configured and constructed CHA to determine if it behaves as expected. Several metrics and evaluation methods are recommended in \cite{abbasian2024foundation}, with additional evaluation techniques explored in \cite{yang2024chatdiet, abbasian2024knowledge, park2024building, abbasian2024empathy}. These evaluations are application-specific, and our framework provides extensive customization capabilities to suit different use cases and requirements. In our future work, we will explore more evaluation techniques.

\noindent
\textbf{User Query Ambiguity:}
Understanding user intentions presents a significant challenge due to query ambiguity, often caused by vague or incomplete information and a lack of necessary external sources connected to openCHA for the specific application. To enhance response accuracy, our future work involves refining openCHA’s ability to clarify user intentions. Key strategies include employing targeted follow-up questions, improving comprehension of the user’s situation, and the precision of responses.

\noindent
\textbf{Scalability:}
We plan to address scalability issues within our system to ensure it can effectively manage increasing loads and accommodate more complex scenarios. This involves enhancing system architecture and resource allocation to support larger data volumes and more simultaneous users without compromising performance. We plan to investigate further how good the openCHA core scalability is in real-world scenarios.


\section{Conclusion}
This paper presented openCHA, an LLM-powered framework designed to empower CHAs in effectively addressing healthcare-related queries through the analysis of input questions, data collection, action execution, and the delivery of personalized responses. We demonstrated the framework's effectiveness using two different demos on general patient health record reporting, objective stress level estimation, and further use cases. Additionally, we discussed the capabilities, limitations, and challenges of openCHA. Our future efforts will focus on improving planning robustness, examining accuracy and evaluation aspects, addressing user query ambiguity, and enhancing the scalability of openCHA.

\bibliographystyle{vancouver}
{\footnotesize
\bibliography{references.bib}}

\section{Appendix 1 - Sample Task Implementation}
We encourage readers and openCHA users to review our documentation page \footnote{\url{https://docs.opencha.com/index.html}} for the latest updates on openCHA.

\begin{figure}
\centering
\includegraphics[width=\textwidth, trim={0cm 0cm 0cm 8cm},clip]{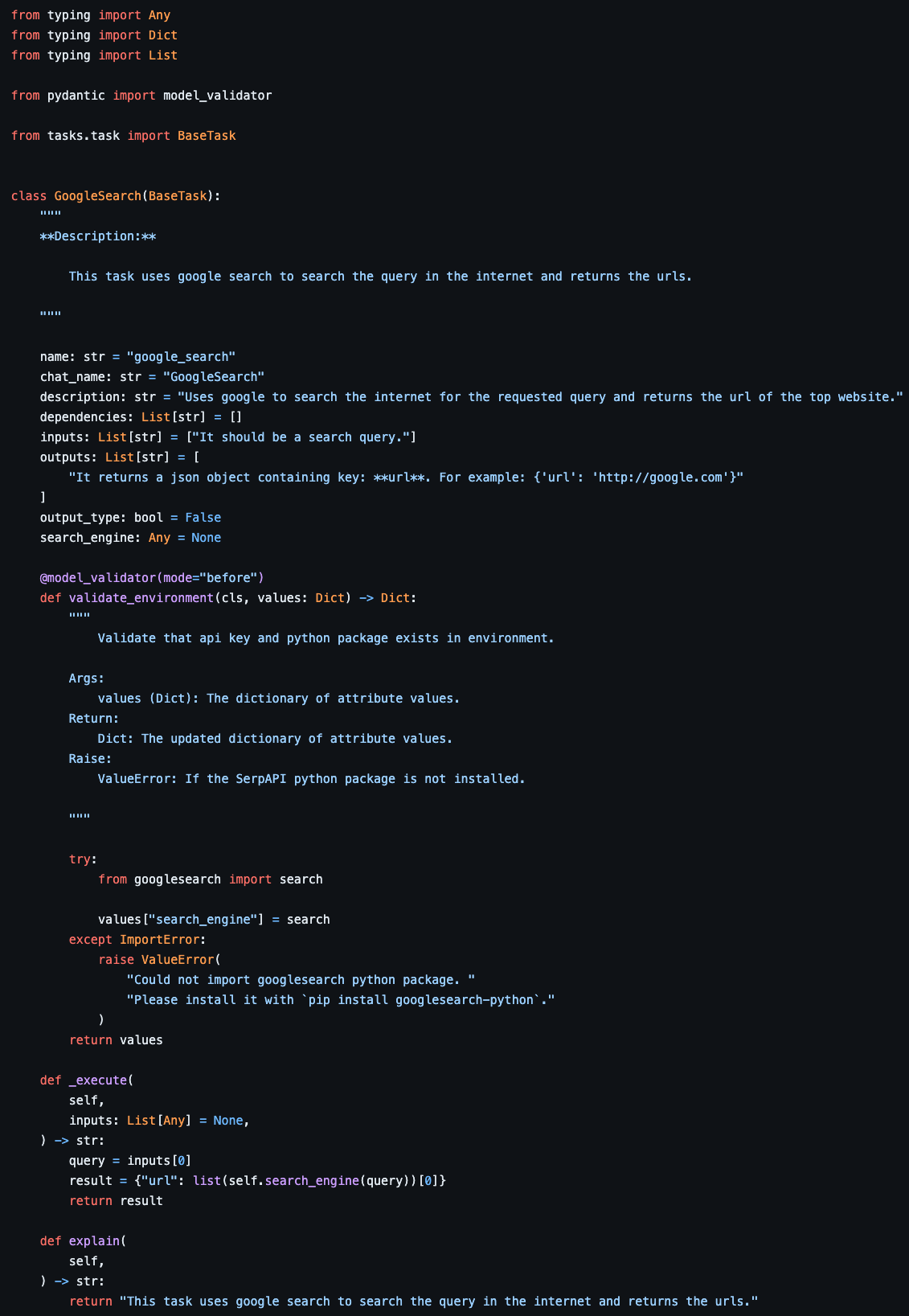}
\caption {A sample implemented task inside openCHA framework. Each task includes prompt parameters which helps our Task Planner LLM understand what types of data are needed and how to provide them to the current task.
\label{fig:task_example}}
\end{figure}

In this appendix, we will detail the curation of tasks within the openCHA framework and their introduction as external sources to the Task Planner. Initially, we will describe the types of tasks that can be added to the framework. Subsequently, we will outline the implementation process for tasks within the framework. Finally, we will present an example of a task that has already been implemented, along with sample prompts associated with it.

Tasks may be implemented locally or as a service. Local implementation entails coding all necessary components on-site to ensure successful task execution, which may involve providing a Python library, a GitHub repository, or a complete task implementation.

Service-based implementation involves hosting services on a server and offering APIs for utilization. In this scenario, the task should invoke these APIs using Python libraries. It is incumbent upon the service provider to maintain the task and supply comprehensive documentation for users, including registration procedures, API key acquisition, and privacy policies.

Figure \ref{fig:task_example} displays an example of a task implementation in openCHA designed to perform a Google search and return the first search result URL. Each task includes configurations to aid the Task Planner LLM in understanding when and how to use this task. The parameters are as follows:

\noindent
\textbf{name}

\textbf{Purpose:} Uniquely identifies the task within the system, and this name will appear in the interface for task selection.

\textbf{Convention:} Generally in lowercase underscore\_case, often derived from the task's functionality for better readability and maintainability.

\textit{Example:} name = "google\_search"

\noindent
\textbf{chat\_name:}

\textbf{Purpose:} Used for referencing the task in user interfaces or chats, particularly for explainability. If a user inquires about the tasks used, these names will be displayed.

\textbf{Convention:} Typically in CamelCase, should be descriptive yet concise.

\textit{Example:} chat\_name = "GoogleSearch"

\noindent
\textbf{description:}

\textbf{Purpose:} Provides an explanation of the task's function. This information helps the Task Planner LLM decide when to deploy the task.

\textbf{Convention:} The description should be comprehensive, outlining task capabilities, prioritization, or specific conditions for its use. For instance, ensuring the Task Planner prioritizes patient data analysis over frequent internet searches can be mentioned here with a low priority setting.

\textit{Example:} description = "Uses Google to search the internet for the requested query and returns the URL of the top website."

\noindent
\textbf{dependencies:}

\textbf{Purpose:} Lists other tasks or services this task depends on, informing the Task Planner LLM of these dependencies during planning.

\textbf{Convention:} An array of task identifiers that match the name attribute of dependent tasks.

\textit{Example:} dependencies = [TaskType.FETCH\_WEATHER\_DATA]

\noindent
\textbf{inputs:}

\textbf{Purpose:} Specifies the inputs required by the task. This parameter is crucial as it guides the Task Planner LLM to provide the appropriate inputs for this task's "\_execute" function.

\textbf{Convention:} An array of strings, each describing a specific input.

\textit{Example:} inputs = ["It should be a search query."]

\noindent
\textbf{outputs:}

\textbf{Purpose:} Details what the task returns, assisting the Task Planner LLM in understanding the return structure for proper handling of the output.

\textbf{Convention:} An array of strings describing each output, with a full description of the return data format.

\textit{Example:} outputs = ["It returns a JSON object containing key: url. For example: {'url': 'http://google.com'}"]

\noindent
\textbf{output\_type}

\textbf{Purpose:} Instructs the Orchestrator on how to handle the output of this task. If output\_type is True, the data is stored in the Datapipe; otherwise, it is directly returned to the Task Planner LLM. This flexibility allows task designers to determine whether their task involves intermediate information for use by other tasks or provides the final answer.

\textbf{Convention:} Boolean values (True or False).

\noindent
\textbf{\_execute function}

\textbf{Purpose:} This function must be implemented correctly. If all other parameters are set accurately, developers can trust that the Task Planner LLM will invoke this function with the appropriate inputs. For instance, as shown in Figure \ref{fig:task_example}, the sole expected input is a search query. The Task Planner LLM will correctly supply the search query as an input to the "\_execute" function, accessible as the first element in the inputs array (since the only entry in the "inputs" configuration pertains to the search query, this array will contain just one item). Utilizing the google\_engine Python library, this function searches for the query online and returns the URL. Note that the result we are returning for the Google search task aligns with what is specified in the "outputs" parameter, demonstrating the coherence between the task setup and its execution.

\section{Appendix 2 - Task Planner Sample Implementation}
We encourage readers and openCHA users to review our documentation page \footnote{\url{https://docs.opencha.com/index.html}} for the latest updates on openCHA.

In this appendix, we elaborate on the implementation of the Tree of Thought prompting technique. We describe how available tasks are introduced to the Tree of Thought, the process of curating and generating the planning prompt, and ultimately, how the final planning results are transformed into executable tasks by the Task Executor.

The core component of the Task Planner is the selection of an appropriate planning technique. Among various techniques using LLMs that have proven practical, we opted for the Tree of Thought approach. 

To effectively implement the Tree of Thought, we structured the planning process into two main sections. Initially, we prepare a comprehensive prompt that includes a list of available tasks, relevant metadata, records of previously performed tasks, conversation history, and user input. This setup ensures the planner is thoroughly informed about the tasks it can call upon, the metadata that should be used or passed along, and the context of past interactions to prevent redundant planning and maintain the continuity of conversations.

\begin{figure}
\centering
\includegraphics[width=\textwidth, trim={0cm 0cm 0cm 0cm},clip]{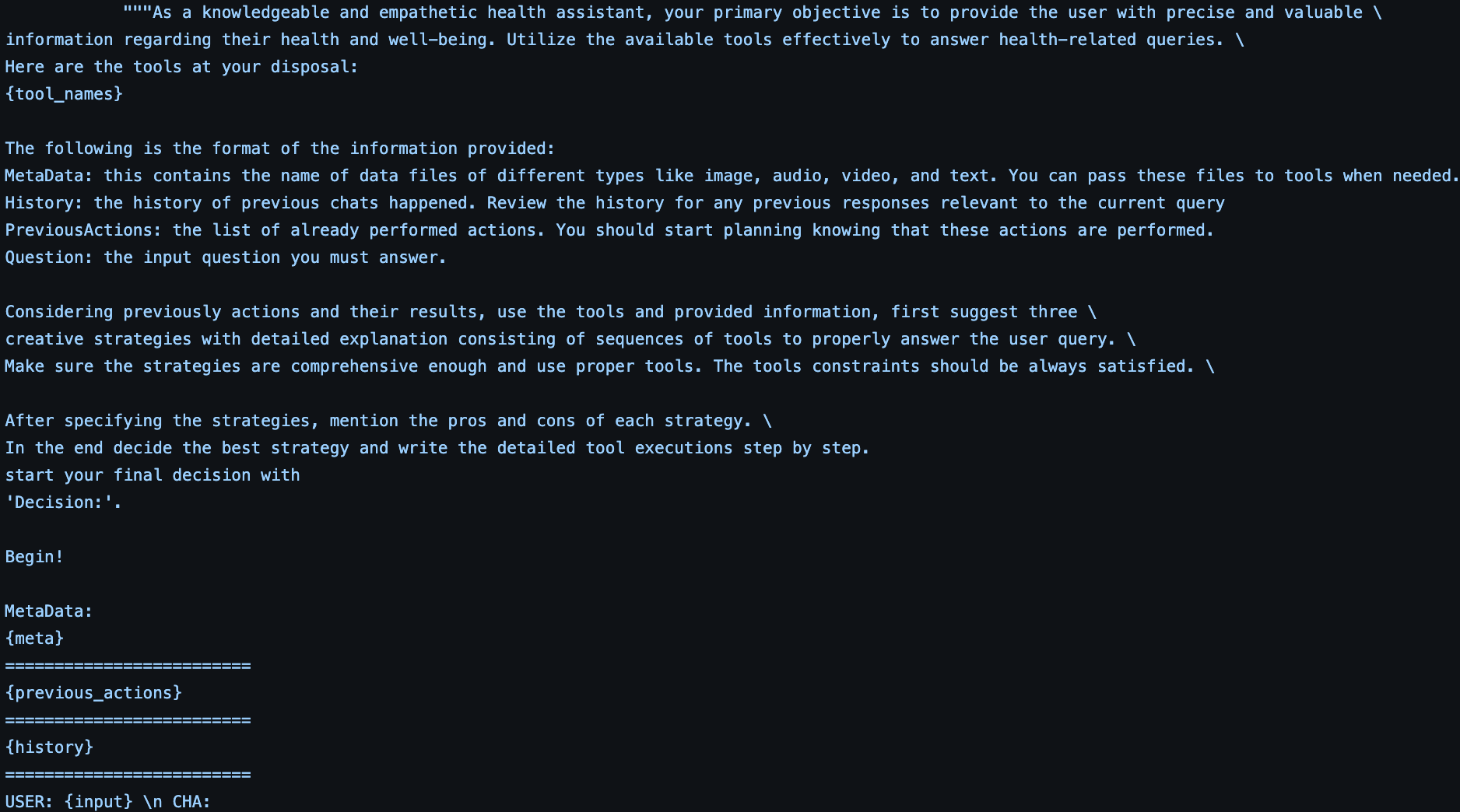}
\caption {The first stage Tree of Thought planning prompt
\label{fig:planner_prompt1}}
\end{figure}

Figure \ref{fig:planner_prompt1} illustrates the Tree of Thought planning prompt. In this phase, the Tree of Thought uses all this gathered information and the user's query to devise three distinct task sequences or strategies. Each strategy is designed to collect the necessary information to address the query efficiently. Next, it is asked to provide the pros and cons of each strategy, ultimately selecting the most suitable one as the final decision. 

In the second stage of the planning process, our objective is to translate the chosen decision into sequences of task functions that the Orchestrator can understand and execute. Within the Orchestrator, we have implemented a function named execute\_task, which serves as an interface to retrieve and execute tasks with the appropriate inputs accurately.

\begin{figure}
\centering
\includegraphics[width=\textwidth, trim={0cm 0cm 0cm 0cm},clip]{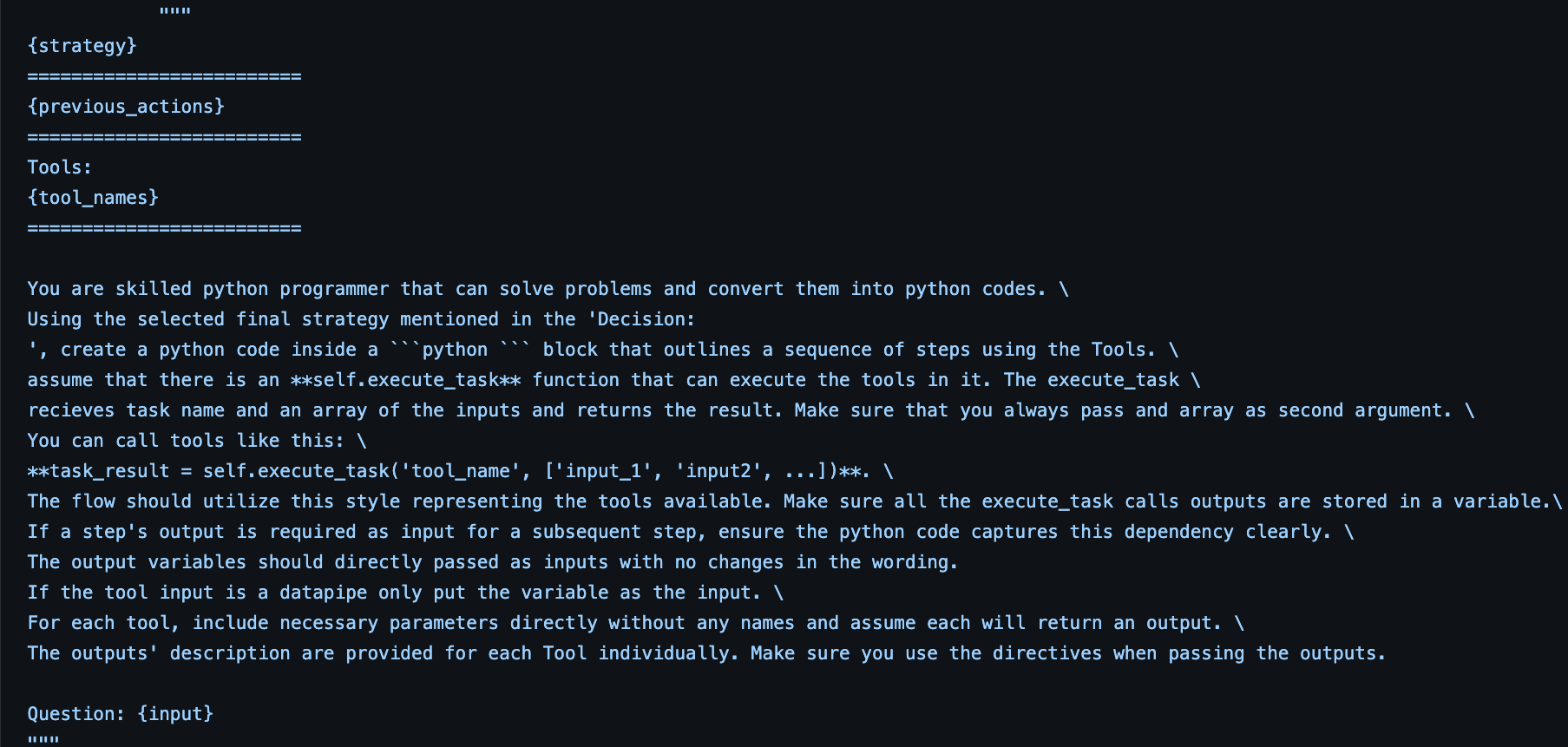}
\caption {The second stage Tree of Thought planning prompt
\label{fig:planner_prompt2}}
\end{figure}

Our second prompt instructs the LLM to invoke this execute\_task function, ensuring that the correct inputs are provided for each task. Figure \ref{fig:planner_prompt2} displays this second prompt, showcasing how we structure these commands within the LLM to streamline task execution.

\section{Appendix 3 - Response Generator implementation}
We encourage readers and openCHA users to review our documentation page \footnote{\url{https://docs.opencha.com/index.html}} for the latest updates on openCHA.

In this appendix, we provide a detailed explanation of the Response Generator's implementation. We begin by distinguishing between the Response Generator and the Task Planner. Following this, we demonstrate how the outcomes of planning and task execution are delivered to the Response Generator. Lastly, we present the prompts employed to ensure that the Response Generator depends on the tasks' responses rather than the internal knowledge of the LLM used for generating responses.

One of the key objectives of the Agentic approach is to create a comprehensive mega prompt that allows an LLM to generate more precise answers by incorporating all necessary information. Our innovation centers on dynamically generating this mega prompt in response to user queries. The Task Planner initiates this process by determining the sequence of tasks it deems necessary to accurately address the current question.

Once planning is complete, the Orchestrator compiles the results of these tasks to create the mega prompt. The composition of this prompt is influenced by the nature of the tasks and their outputs. For instance, if a task's output\_type is set to true, indicating that the result should be stored in the Datapipe for subsequent analysis task use, the Orchestrator will omit this result from the prompt. The focus in this case is on the analysis's outcome rather than intermediate results. Similarly, for tasks that produce metadata, such as an image, the metadata itself isn't passed directly to the Response Generator. Instead, descriptive information about the metadata, like details of a plot image and relevant data insights, are provided to the Response Generator, indicating that this information accompanies the image.

\begin{figure}
\centering
\includegraphics[width=\textwidth, trim={0cm 0cm 0cm 0cm},clip]{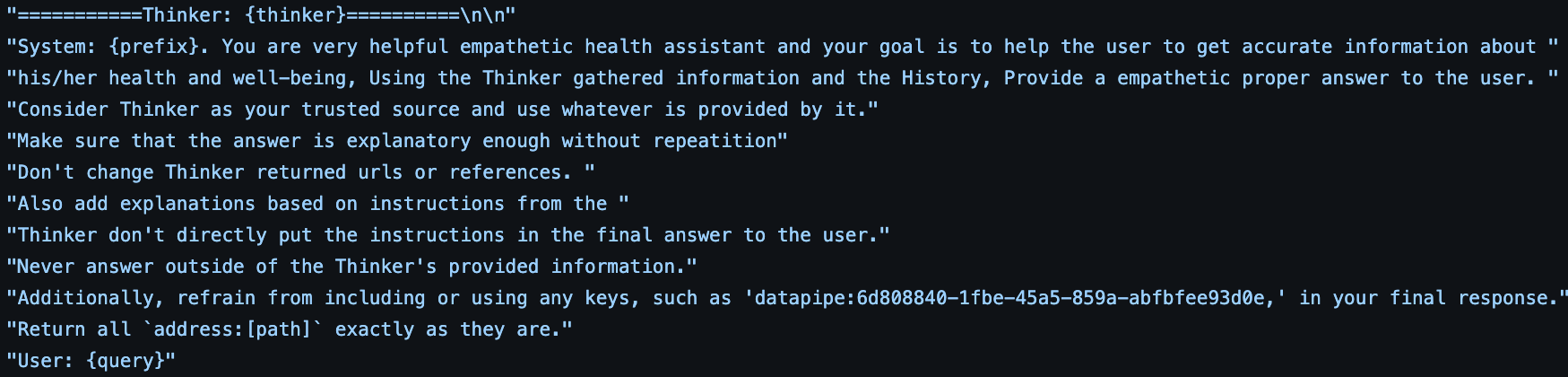}
\caption {The Response Generator prompt.
\label{fig:response_generator}}
\end{figure}

Figure \ref{fig:response_generator} illustrates the prompt provided to the Response Generator LLM. Initially, the Orchestrator formats answers derived from the task results, which we refer to as the "thinker" part of the prompt. The LLM is then instructed to rely exclusively on this thinker prompt, disregarding its internal knowledge. Additionally, we include guidelines to help the LLM understand specific system implementations, such as Datapipe access keys or metadata details.

In the openCHA framework, this prompt is customizable. A "prefix" feature allows developers to add information to better align the response generator with their specific task requirements.

\section{Appendix 4 - Sample Prompts}
In this Appendix, we demonstrate the generated prompt for a simple example involving two tasks: Internet search and webpage information extraction. The user's query, "How to improve my sleep," is first directed to the Task Planner. Table \ref{tab:planner_prompt1} displays the initial planning stage where the Tree of Thought Planner is requested to develop three strategies, evaluate their pros and cons, and choose one to execute. Table \ref{tab:planner_prompt2} presents the prompt for the second planning stage, where the LLM is instructed to generate Python code that the Orchestrator can execute based on the chosen strategy, along with the necessary inputs. Table \ref{tab:planner_prompt3} and \ref{tab:planner_prompt4} provides an example of the generated code, illustrating how the LLM correctly inputs the query "tips to improve sleep" for the google\_search task. Table \ref{tab:response_generator_prompt} shows how openCHA creates a prompt to assist the Response Generator in utilizing the tasks' results to respond to the user's query effectively.

Additionally, Table \ref{tab:task_prompt} provides an example of a prompt for a task that has multiple inputs and outputs, with the result being stored in the Data Pipe. Table \ref{tab:task_prompt2} illustrates the process of passing the Data Pipe key between tasks. In this setup, each task returns a Data Pipe key, and within the generated code, the output from one task is appropriately passed to the next task as its input. The execute\_task function manages access to the Data Pipe when a key is provided, ensuring that the results of the tasks are not directly transmitted to the LLM. 

\begin{table}
    \centering
\caption{Tree of Thought Planning first planning stage.}
\label{tab:planner_prompt1}
    \begin{tabular}{|>{\raggedright\arraybackslash}p{1\linewidth}|} \hline 
        As a knowledgeable and empathetic health assistant, your primary objective is to provide the user with precise and valuable information regarding their health and well-being. Utilize the available tools effectively to answer health-related queries. Here are the tools at your disposal:

-----------------------------------

**google\_search**: 
Uses google to search the internet for the requested query and returns the url of the top website.
This tool have the following outputs:

It returns a json object containing key: **url**. For example: {'url': 'http://google.com'}

-----------------------------------

**extract\_text**: Extract all the text on the current webpage
This tool have the following outputs:

An string containing the text of the scraped webpage.

-----------------------------------

The following is the format of the information provided:
MetaData: This contains the names of data files of different types, such as images, audio, video, and text. You can pass these files to tools when needed.

History: The history of previous chats happened. Review the history of any previous responses relevant to the current query.

PreviousActions: the list of actions that have already been performed. You should start planning, knowing that these actions are performed.

Question: The input question that you must answer.

Considering previously actions and their results, use the tools and provided information, first suggest three creative strategies with detailed explanation consisting of sequences of tools to properly answer the user query. Make sure the strategies are comprehensive enough and use proper tools. The tools constraints should be always satisfied. After specifying the strategies, mention the pros and cons of each strategy. In the end, decide the best strategy and write the detailed tool executions step by step. start your final decision with

'Decision:'.

Begin!

=========================

USER: How to improve my sleep? 

CHA:

\\ \hline 
    \end{tabular}

\end{table}

\begin{table}
    \centering
\caption{Tree of Thought Planning second planning stage: generating the task execution code part 1.}
\label{tab:planner_prompt2}
    \begin{tabular}{|>{\raggedright\arraybackslash}p{1\linewidth}|} \hline 
        Decision:

I will go with Strategy 1 as it provides the most recent and relevant information available on the internet, which is crucial for improving sleep. 

Now, let's proceed with the detailed tool executions for Strategy 1:

1. Use the "google\_search" tool to find the top websites with tips to improve sleep.

2. Once we have the top website, we can use the "extract\_text" tool to extract the relevant information about improving sleep from the webpage.

Let's start with step 1.

=========================

Tools:

-----------------------------------

**google\_search**: Uses google to search the internet for the requested query and returns the url of the top website.

  The input to this tool should be a list of data representing:
  
   1-It should be a search query.

   This tool will return the following data:
   
- It returns a json object containing key: **url**. For example: {'url': 'http://google.com'}

-----------------------------------

**extract\_text**: Extract all the text on the current webpage

  The input to this tool should be a list of data representing:
  
   1-url to extract the text from. It requires links which is gathered from other tools. Never provide urls on your own.

   This tool will return the following data:
   
- An string containing the text of the scraped webpage.

-----------------------------------
\\ \hline
    \end{tabular}

\end{table}

\begin{table}
    \centering
\caption{Tree of Thought Planning second planning stage: generating the task execution code part 2.}
\label{tab:planner_prompt3}
    \begin{tabular}{|>{\raggedright\arraybackslash}p{1\linewidth}|} \hline 

=========================

You are a skilled Python programmer who can solve problems and convert them into Python codes. Using the selected final strategy mentioned in the 'Decision:
', create a python code inside a ```python ``` block that outlines a sequence of steps using the Tools. Assume that there is a **self.execute\_task** function that can execute the tools in it. The execute\_task receives the task name and an array of the inputs and returns the result. Make sure that you always pass an array as a second argument. You can call tools like this: **task\_result = self.execute\_task('tool\_name', ['input1', 'input2', ...])**. The flow should utilize this style to represent the tools available. Make sure all the execute\_task calls outputs are stored in a variable. If a step's output is required as input for a subsequent step, ensure the Python code captures this dependency clearly. The output variables should be directly passed as inputs with no changes in the wording.
If the tool input is a datapipe, only put the variable as the input. For each tool, include necessary parameters directly without any names and assume each will return an output. The outputs' description are provided for each tool individually. Make sure you use the directives when passing the outputs.

Question: How to improve my sleep?\\ \hline
    \end{tabular}

\end{table}

\begin{table}
    \centering
\caption{Sample generated code for task execution.}
\label{tab:planner_prompt4}
    \begin{tabular}{|>{\raggedright\arraybackslash}p{1\linewidth}|} \hline 

\# Step 1: Use google\_search to find the top websites with tips to improve sleep.

search\_query = "tips to improve sleep"
search\_result = self.execute\_task('google\_search', [search\_query])

\# Step 2: Use extract\_text to extract the relevant information about improving sleep from the webpage.
url = search\_result['url']

sleep\_tips\_text = self.execute\_task('extract\_text', [url])\\ \hline
    \end{tabular}

\end{table}

\begin{table}
    \centering
\caption{Response generator sample prompt.}
\label{tab:response_generator_prompt}
    \begin{tabular}{|>{\raggedright\arraybackslash}p{1\linewidth}|} \hline 

===========Thinker: 

MetaData: 

History: 

------------------

google\_search: ['tips to improve sleep']

{'url': 'https://www.mayoclinic.org/healthy-lifestyle/adult-health/in-depth/sleep/art-20048379'}

------------------

------------------

extract\_text: ['https://www.mayoclinic.org/healthy-lifestyle/adult-health/in-depth/sleep/art-20048379']

Sleep tips: 6 steps to better sleep - Mayo Clinic This content does not have an English version. This content does not have an Arabic version. Skip to content Care at 
 Mayo Clinic Patient-Centered Care About Mayo... (we cut the text to shorten the table)
 
------------------

==========

System: . You are a very helpful, empathetic health assistant, and your goal is to help the user get accurate information about his/her health and well-being; using the Thinker gathered information and the History, Provide an empathetic, proper answer to the user. Consider Thinker as your trusted source, and use whatever it provides. Make sure that the answer is explanatory enough. Don't change Thinker returned URLs or references. Also, add explanations based on instructions from the Thinker. Don't directly put the instructions in the final answer to the user. Never answer outside of the Thinker's provided information. Additionally, refrain from including or using any keys, such as 'datapipe:6d808840-1fbe-45a5-859a-abfbfee93d0e,' in your final response. Return all `address:[path]` exactly as they are.User: How to improve my sleep?\\ \hline
    \end{tabular}

\end{table}

\begin{table}
    \centering
\caption{Task prompt with Data Pipe.}
\label{tab:task_prompt}
    \begin{tabular}{|>{\raggedright\arraybackslash}p{1\linewidth}|} \hline 

-----------------------------------

**affect\_ppg\_get**: Returns the ppg data for a specific patient over a date or a period (if two dates are provided). This will return the detailed raw data and store it in the Data Pipe.

  The input to this tool should be a list of data representing:
  
   1-user ID in string. It can be referred to as user, patient, individual, etc. Start with 'par\_' followed by a number (e.g., 'par\_1').

  2-start date of the sleep data in a string with the following format: `\%Y-\%m-\%d.`

  3-end date of the sleep data in a string with the following format: `\%Y-\%m-\%d.` If there is no end date, the value should be an empty string (i.e., '')

   This tool will return the following data:
   
- returns an array of JSON objects which contains the following keys:

**date (in milliseconds)**: epoch format

**ppg**: is the ppg value.

**hr (in beats per minute)**: is the heart rate of the patient.

 The result will be stored in the Data Pipe.
 
-----------------------------------\\ \hline
    \end{tabular}

\end{table}

\begin{table}
    \centering
\caption{Sample code generation when a task's result is stored in the Data Pipe.}
\label{tab:task_prompt2}
    \begin{tabular}{|>{\raggedright\arraybackslash}p{1\linewidth}|} \hline 

\# Step 1: Get PPG data for patient 5 for the entire month of August 2020

ppg\_data\_result = self.execute\_task('affect\_ppg\_get', ['par\_5', '2020-08-01', '2020-08-31'])

\# Step 2: Analyze the HRV parameters from the obtained PPG data

hrv\_analysis\_result = self.execute\_task('affect\_ppg\_analysis', [ppg\_data\_result])

\# Step 3: Estimate the stress level for patient 5 during August 2020 using the HRV analysis results

stress\_level\_result = self.execute\_task('affect\_stress\_analysis', [hrv\_analysis\_result])\\ \hline
    \end{tabular}

\end{table}

\end{document}